\documentclass[preprint,review,10pt]{elsarticle}

\usepackage[a4paper,left=2.5cm,right=2.5cm,top=3cm,bottom=3cm]{geometry}
\usepackage[parfill]{parskip}  

\usepackage[english]{babel}
\usepackage{lmodern}            
\usepackage{bm}                 

\usepackage{amssymb, amstext, amsthm, mathtools}

\newcommand{\xvec}{\boldsymbol{x}}

\usepackage{graphicx}           
\usepackage{subcaption}         
\usepackage{tikz}               
\usetikzlibrary{shapes.geometric, arrows}

\tikzstyle{startstop} = [ellipse, minimum width=3cm, minimum height=1cm,
    text centered, draw=black, line width=1.5pt, fill=red!15]
\tikzstyle{io} = [trapezium, trapezium left angle=70, trapezium right angle=110,
    minimum width=2.5cm, minimum height=1cm, text centered, draw=black, line width=1.5pt, fill=blue!15]
\tikzstyle{process} = [rectangle, rounded corners=6pt, minimum width=4cm, minimum height=1cm,
    text centered, draw=black, line width=1.5pt, fill=teal!15]
\tikzstyle{decision} = [diamond, aspect=2, text centered, draw=black, line width=1.5pt, fill=orange!20]
\tikzstyle{arrow} = [thick,->,>=stealth, line width=2.5pt]

\usepackage{booktabs, tabu, tabularx, float}
\usepackage{enumitem}           

\usepackage[ruled,vlined]{algorithm2e}

\usepackage[bookmarks=true,colorlinks=true,linkcolor=blue,citecolor=red]{hyperref}
\usepackage[nameinlink,capitalize]{cleveref}
\makeatletter
\AtBeginDocument{\def\@citecolor{red}}
\def\ps@pprintTitle{%
  \let\@oddhead\@empty
  \let\@evenhead\@empty
  \def\@oddfoot{}%
  \let\@evenfoot\@oddfoot}
\makeatother

\begin{document}


\begin{frontmatter}

\title{Adaptive Hard–Soft Physics-Informed Neural Networks for Robust Boundary-Constrained PDE Solving}



\author[1,2]{Duc Tien Nguyen\fnref{equal}}
\author[3]{Trinh Minh Tuan\fnref{equal}}
\author[4]{Nguyen Duc Manh}
\author[1,2]{Vu Linh Nguyen}
\author[3]{Dinh Gia Ninh}


\fntext[equal]{These authors contributed equally to this work.}

\affiliation[1]{%
 organization={College of Engineering and Computer Science, VinUniversity}, 
city={Hanoi},
country={Vietnam}
}

\affiliation[2]{%
organization={Center for AI Research, VinUniversity}, 
city={Hanoi},
country={Vietnam}
}

\affiliation[3]{%
 organization={Group of Materials and Structures, School of Mechanical Engineering, Hanoi University of Science and Technology}, 
city={Hanoi},
country={Vietnam}
}

\affiliation[4]{%
  organization={Department of Mathematics and Informatics, Hanoi University of Science and Technology}, 
  city={Hanoi},
  country={Vietnam}
}

    

\begin{abstract}
Physics-informed neural networks (PINNs) provide an effective way to solve partial differential equations (PDEs) by embedding physical principles into the learning process. However, the conventional PINN formulation, in which all constraints are imposed as soft penalty terms within a composite loss, often exhibits slow convergence, sensitivity to loss weight scaling, and inaccurate boundary enforcement due to poor conditioning of the optimization landscape. To address these limitations, this study proposes a unified hard--soft physics--informed neural network (HSPINN) with adaptive loss weighting. In this framework, Dirichlet and periodic boundary conditions are enforced exactly by construction through analytical or polynomial lifting, masking functions, and periodic feature mappings, while the governing PDE residuals, Neumann fluxes, and initial conditions are treated as soft constraints. An inverse-share softmax strategy dynamically balances the relative importance of individual loss components during training, eliminating manual penalty tuning and improving gradient stability. This formulation ensures boundary admissibility throughout optimization and enhances convergence efficiency and numerical robustness. Applications to representative elliptic (Poisson), parabolic (Burgers), and hyperbolic (convection with periodic boundaries) problems demonstrate that HSPINN consistently achieves faster convergence, higher accuracy, and greater stability than conventional PINNs, establishing a general and scalable foundation for physics constrained deep learning across science and technology.
\end{abstract}


\end{frontmatter}

\section{Introduction}

Partial differential equations are widely regarded as fundamental models for describing physical, biological, and engineering systems \citep{nguyen2025inverse, nguyen2025modeling}. They are encountered in heat transfer, fluid flow, elasticity, electromagnetism, and chemical diffusion. Conventional numerical techniques, including finite difference, finite element, and spectral methods, have reached a high degree of maturity and accuracy \citep{soga2024mathematical, hughes2012finite}. Nevertheless, mesh generation, stabilization procedures, and domain discretization are required. When geometries are complex, solution features are multiscale, or coupling with heterogeneous data is necessary, these prerequisites can become computationally expensive and practically restrictive.

Recently, physics-informed neural networks have emerged as an alternative computational paradigm~\citep{akrivis2025runge, biswas2022error, nguyen2025highly, tan2024utilizing}. Within this framework, the solution of a PDE is approximated by a neural function that is trained through minimization of a composite loss embedding the governing operator together with boundary and initial conditions. Because spatial and temporal derivatives are obtained by automatic differentiation (AD) and no explicit mesh is required, adaptation to irregular domains and the incorporation of observational data can be achieved naturally. As a result, a strong synergy between data and physics has been established, and PINNs have been widely adopted as flexible neural PDE solvers.

Despite these advantages, several limitations have been identified in the conventional PINN formulation, particularly in the treatment of boundary and interface conditions. In the canonical approach, all constraints including the PDE residual, Dirichlet data, Neumann data, and initial conditions are imposed as soft penalties within a weighted objective. The magnitudes of these partial losses often differ by several orders, which leads to loss imbalance, unstable optimization dynamics, and poor boundary adherence \citep{cuomo2022scientific, wang2021understanding, krishnapriyan2021characterizing}. Although adaptive reweighting, curriculum scheduling, and domain decomposition strategies such as XPINNs have been proposed \citep{jagtap2020extended}, exact satisfaction of boundary conditions has not been guaranteed and sensitivity to manually chosen penalty weights has remained.

To address these difficulties, hard constraint formulations have been proposed in which certain boundary conditions are satisfied by construction rather than by penalty \citep{lu2021physics}. Distance based parameterizations and augmented Lagrangian ideas were introduced in \texttt{DeepXDE} \citep{lu2021deepxde}, followed by distance function and signed distance approaches designed to enforce exact Dirichlet satisfaction \citep{song2022hardpinn, karumuri2023hardbc}. These developments have improved boundary accuracy, but the emphasis has typically been placed on Dirichlet conditions, while Neumann data have remained weakly enforced and training has continued to rely on delicate loss weight choices. Moreover, the evaluation of distance functions and the use of geometry aligned spectral decompositions have increased computational cost and architectural complexity.  

In parallel, several studies have highlighted the difficulty of enforcing periodic boundaries within the conventional PINN framework \citep{awwal2023generalized, hao2024structure}. When periodicity is treated through soft penalty terms, the network often fails to reproduce smooth transitions across opposite boundaries, resulting in phase discontinuities, loss imbalance, and degraded convergence. These issues arise because periodic constraints couple both the solution and its derivatives at distant points in the domain, which are not naturally captured by localized residual sampling. Although specialized formulations have been proposed to improve periodic \citep{cho2024parameterized}, their applicability has remained confined to fully periodic domains and has not been extended to mixed or heterogeneous boundary configurations. Consequently, a unified and computationally efficient formulation capable of handling Dirichlet, Neumann, and periodic boundaries within a single learning framework has not yet been fully established.

In this study, a hard-soft physics-informed neural network  framework is proposed to overcome these limitations. Within this formulation, the network representation is constructed such that Dirichlet conditions are satisfied exactly by design through analytical or polynomial lifting together with masking functions, thereby eliminating the need for Dirichlet penalty terms and restricting the hypothesis space to boundary admissible functions. The remaining physical relations, including the governing PDE, Neumann fluxes, and initial conditions, are treated as soft constraints through residual minimization. Periodic boundaries are incorporated naturally within the same architecture, ensuring intrinsic periodicity without additional penalty terms and enabling consistent treatment alongside other boundary types. To stabilize multi term optimization without manual tuning, an inverse-share softmax weighting scheme is employed to adaptively balance the contributions of the individual residuals. Through this mechanism, scale disparities among loss components are mitigated, convergence is accelerated, and gradient stability is improved.

The proposed framework is demonstrated on three representative classes of PDEs: an elliptic Poisson problem with mixed Dirichlet and Neumann boundaries, a parabolic viscous Burgers problem with prescribed Dirichlet and initial conditions, and a hyperbolic convection problem with periodic boundaries. In all cases, faster convergence, improved accuracy, and stronger boundary adherence are achieved compared with conventional fully soft PINNs and representative hard constraint variants, without any increase in model complexity. The major contributions of this paper are summarized as follows:
\begin{itemize}
    \item A unified neural parameterization is introduced in which Dirichlet boundaries are enforced exactly through lifting and masking, yielding boundary admissible hypotheses without penalty terms.
    \item A single formulation is established that consistently handles Dirichlet, Neumann, and periodic boundary conditions within the same architecture, enabling general applicability across different PDE types.
    \item An adaptive inverse-share softmax weighting strategy is developed to balance multiple residual components automatically, which improves conditioning, accelerates convergence, and removes manual weight tuning.
\end{itemize}

These elements establish a general and efficient framework for learning solutions of PDEs with accurate boundary treatment and robust optimization behavior, and are expected to be broadly applicable to coupled transport, diffusion, and wave systems in scientific and engineering contexts.

\section{Formulation}
In this section, the mathematical and algorithmic formulation of the proposed HSPINN framework is presented. The governing partial differential equations and their associated boundary and initial conditions are first defined, establishing the general problem setting. The neural parameterization that enables exact Dirichlet and periodic enforcement through lifting and masking is then introduced, followed by the construction of residuals and the composite loss function used for training. An adaptive weighting strategy is described to balance the contributions of the individual residual terms, and the overall optimization procedure is detailed, including stochastic pretraining and quasi-Newton refinement.

\subsection{Governing Equations}
\label{sec:gov_eqs}

The physical systems considered in this study are modeled by partial differential equations on bounded spatial domains 
$\Omega \subset \mathbb{R}^d$ with boundary $\partial\Omega$. 
For boundary-value problems, the boundary is partitioned into Dirichlet and Neumann subsets, denoted by 
$\Gamma_D$ and $\Gamma_N$, respectively, such that 
$\partial\Omega = \Gamma_D \cup \Gamma_N$ and 
$\Gamma_D \cap \Gamma_N = \varnothing$. 
For time-dependent problems, the spatial domain is extended over a finite time interval $[0,T]$, forming the space--time domain 
$Q_T = \Omega \times (0,T]$.

The unknown field $u(\mathbf{x},t)$ represents a scalar physical quantity (e.g., potential or velocity component). 
Its evolution is governed by a differential operator $\mathcal{N}[\cdot]$, which may be linear or nonlinear and may involve first- or second-order derivatives. 
The strong form of the generic problem is written as
\begin{equation}
    \mathcal{N}[u](\mathbf{x},t) = f(\mathbf{x},t),
    \qquad (\mathbf{x},t) \in Q_T,
\end{equation}
where $f$ is a prescribed source. 
Well-posedness is completed by boundary and initial data:
\begin{equation}
\begin{aligned}
    u(\mathbf{x},t) &= g_D(\mathbf{x},t), 
    && \mathbf{x} \in \Gamma_D,\ t \in (0,T], \\[2pt]
    \mathbf{n} \cdot \nabla u(\mathbf{x},t) &= g_N(\mathbf{x},t),
    && \mathbf{x} \in \Gamma_N,\ t \in (0,T], \\[2pt]
    u(\mathbf{x},0) &= u_0(\mathbf{x}), 
    && \mathbf{x} \in \Omega,
\end{aligned}
\end{equation}
with $\mathbf{n}$ the outward unit normal and compatibility assumed on $\Gamma_D$ at $t=0$. 
Periodic boundaries, when present, identify opposite faces of $\partial\Omega$ and are handled architecturally (see \cref{sec:results}).

Three representative PDEs are considered to span the elliptic, parabolic, and hyperbolic classes:
\begin{itemize}
    \item \emph{Elliptic (Poisson):}
    \begin{equation}
        -\,\Delta u = f \quad \text{in } \Omega,
    \end{equation}
    posed on the unit square with a mixed Dirichlet--Neumann partition $(\Gamma_D,\Gamma_N)$.

    \item \emph{Parabolic (viscous Burgers):}
    \begin{equation}
        u_t + u\,u_x - \nu\,u_{xx} = f \quad \text{in } (x,t)\in(0,1)\times(0,T],
    \end{equation}
    with nonzero Dirichlet data $u(0,t)=u(1,t)=g_D(t)$ and an initial condition $u(x,0)=u_0(x)$.

    \item \emph{Hyperbolic (pure convection):}
    \begin{equation}
        u_t + \beta\,u_x = f \quad \text{in } (x,t)\in[0,2\pi]\times(0,T],
    \end{equation}
    with \emph{periodic} boundary conditions in $x$ and a soft initial condition $u(x,0)=u_0(x)$.
\end{itemize}

In the proposed framework, the PDE and soft constraints are enforced via \emph{pointwise residuals} evaluated by AD. 
For a candidate $u(\mathbf{x},t)$, the residuals read
\begin{equation}
\begin{aligned}
    \mathcal{R}_{\mathrm{int}}(\mathbf{x},t) &= 
    \mathcal{N}[u](\mathbf{x},t) - f(\mathbf{x},t), \\[3pt]
    \mathcal{R}_{\mathrm{bc}}(\mathbf{x},t) &= 
    \mathbf{n}\!\cdot\!\nabla u(\mathbf{x},t) - g_N(\mathbf{x},t), \\[3pt]
    \mathcal{R}_{\mathrm{ic}}(\mathbf{x}) &=
    u(\mathbf{x},0) - u_0(\mathbf{x}).
\end{aligned}
\end{equation}
Unlike conventional PINNs \citep{raissi2019physics}, Dirichlet constraints are \emph{not} included as penalties because they are satisfied \emph{exactly} by construction through an analytic lifting and masking strategy (cf.\ \cref{sec:param}). 
Periodic conditions for the convection case are encoded in the network features, yielding an intrinsically periodic representation and eliminating explicit boundary penalties.

This formulation unifies the treatment of elliptic, parabolic, and hyperbolic problems under a single learning framework: 
Dirichlet conditions are embedded analytically, while Neumann and initial constraints, together with the PDE, are enforced softly through residual minimization. 
The next section details the neural parameterization that enables exact Dirichlet satisfaction and periodic enforcement, forming the core of the proposed hard--soft strategy.

\subsection{Neural Parameterization with Hard--Soft Constraints}
\label{sec:param}
This section introduces the neural parameterization used to embed hard and soft constraints in the proposed framework. Conventional soft PINNs are first revisited to highlight the limitations of penalty-based training, followed by the hard–soft formulation that enforces Dirichlet boundaries exactly while treating the remaining physics as soft residuals.
\subsubsection{Conventional PINNs}

PINNs approximate the solution $u(\mathbf{x},t)$ of a PDE by a neural function $u_\theta(\mathbf{x},t)$ parameterized by weights~$\theta$ \citep{raissi2019physics}. 
Model training is achieved by minimizing a composite loss that penalizes violations of the governing equations and all associated constraints. 
For the general problem described in \cref{sec:gov_eqs}, the conventional, fully soft formulation (hereafter \emph{SPINN}, for soft PINN) is expressed as

\begin{equation}
    \mathcal{L}_{\text{PINN}}(\theta)
    = \lambda_{\text{int}} \mathcal{L}_{\text{int}}
    + \lambda_{D}\mathcal{L}_{D}
    + \lambda_{N}\mathcal{L}_{N}
    + \lambda_{\text{ic}}\mathcal{L}_{\text{ic}},
    \label{eq:pinn_loss_conventional}
\end{equation}
where
\begin{equation}
\begin{aligned}
    \mathcal{L}_{\text{int}}(\theta)
        &= \mathbb{E}_{(\mathbf{x},t)\sim\Omega}
           \!\left[(\mathcal{N}[u_\theta](\mathbf{x},t)-f(\mathbf{x},t))^2\right], \\[3pt]
    \mathcal{L}_{D}(\theta)
        &= \mathbb{E}_{(\mathbf{x},t)\sim\Gamma_D}
           \!\left[(u_\theta(\mathbf{x},t)-g_D(\mathbf{x},t))^2\right], \\[3pt]
    \mathcal{L}_{N}(\theta)
        &= \mathbb{E}_{(\mathbf{x},t)\sim\Gamma_N}
           \!\left[(\mathbf{n}\!\cdot\!\nabla u_\theta(\mathbf{x},t)-g_N(\mathbf{x},t))^2\right], \\[3pt]
    \mathcal{L}_{\text{ic}}(\theta)
        &= \mathbb{E}_{\mathbf{x}\sim\Omega}
           \!\left[(u_\theta(\mathbf{x},0)-u_0(\mathbf{x}))^2\right].
\end{aligned}
\label{eq:soft_losses_full}
\end{equation}

The interior term $\mathcal{L}_{\text{int}}$ enforces the PDE in the domain, 
$\mathcal{L}_{D}$ and $\mathcal{L}_{N}$ penalize deviations from Dirichlet and Neumann boundary conditions, respectively, 
and $\mathcal{L}_{\text{ic}}$ ensures consistency with the initial condition.  
Each constraint is treated as a \emph{soft penalty}, enforced only in an average sense through optimization.

Although conceptually simple, this fully soft formulation suffers from several well-known limitations.  
The relative magnitudes of the losses may differ drastically, leading to unstable or biased training.  
If $\lambda_{D}$ is small, the boundary values may be violated even when the PDE residual is small;  
if $\lambda_{D}$ is large, the optimization becomes dominated by boundary penalties, causing slow convergence or stiffness in the gradient updates.  
This conflict between physical and boundary losses, often termed \emph{loss imbalance}, produces highly non-convex energy landscapes and large gradient disparities across terms \citep{wang2021understanding,krishnapriyan2021characterizing}.  
Consequently, convergence strongly depends on manual tuning of $\lambda_i$ and learning-rate schedules.  
Adaptive weighting \citep{mcclenny2020self,wang2022respecting}, domain decomposition \citep{jagtap2020extended}, and curriculum training \citep{yang2022learning} alleviate but do not remove these issues, because all conditions remain soft penalties.  
A more principled approach is to enforce the exactly known constraints directly in the network architecture.

\subsubsection{Proposed Hard--Soft Constraint Formulation} \label{sec:soft}

To overcome the limitations of fully penalty-based training, a hybrid \emph{hard--soft} enforcement strategy is adopted.  
In this framework, Dirichlet and periodic boundary conditions are embedded into the neural representation so that they are satisfied exactly, 
while the PDE residual, Neumann flux, and initial condition are still enforced softly through residual minimization.  
This distinction eliminates the need for a Dirichlet penalty term and yields a better-conditioned loss landscape.

The network output is defined as
\begin{equation}
    u_\theta(\mathbf{x},t)
    = h_D(\mathbf{x},t) + m_D(\mathbf{x})\,v_\theta(\mathbf{x},t),
    \label{eq:hardsoft_representation}
\end{equation}
where $h_D(\mathbf{x},t)$ is a lifting function satisfying $h_D|_{\Gamma_D}=g_D$, 
$m_D(\mathbf{x})$ is a smooth mask function with $m_D|_{\Gamma_D}=0$ and $m_D>0$ in $\Omega$, 
and $v_\theta(\mathbf{x},t)$ is a trainable neural function representing the free (interior) component.  
By construction, $u_\theta|_{\Gamma_D}=h_D|_{\Gamma_D}=g_D$ holds for any $\theta$, so Dirichlet boundaries are satisfied identically.

For homogeneous Dirichlet conditions, $h_D\equiv0$.  
For non-homogeneous data, $h_D$ is chosen as the analytical or prescribed boundary trace.  
On rectangular domains, a simple polynomial mask,
\begin{equation}
    m_D(x,y)=x(1-x)\,y(1-y),
\end{equation}
vanishes exactly on all edges; in one dimension $m_D(x)=x(1-x)$.  
Smooth alternatives such as $\tanh$-based products can be employed to maintain differentiability near the boundary.  
Periodic boundaries are handled by constructing periodic feature maps, for example 
$\phi(x,t)=[\sin(2\pi x/L),\,\cos(2\pi x/L),\,t]$, so that $u_\theta(\phi(x,t))$ is periodic over $x\in[0,L]$.

Once the exact boundaries are embedded, only the remaining physics are treated as soft constraints.  
Defining the residuals
\begin{equation}
\begin{aligned}
    \mathcal{R}_{\text{int}}(\mathbf{x},t)
        &= \mathcal{N}[u_\theta](\mathbf{x},t) - f(\mathbf{x},t),\\
    \mathcal{R}_{N}(\mathbf{x},t)
        &= \mathbf{n}\!\cdot\!\nabla u_\theta(\mathbf{x},t) - g_N(\mathbf{x},t),\\
    \mathcal{R}_{\text{ic}}(\mathbf{x})
        &= u_\theta(\mathbf{x},0) - u_0(\mathbf{x}),
\end{aligned}
\end{equation}
the total loss becomes
\begin{equation}
    \mathcal{L}_{\text{HS}}(\theta)
    = \lambda_{\text{int}}\mathbb{E}_{\Omega}[\mathcal{R}_{\text{int}}^2]
    + \lambda_{N}\mathbb{E}_{\Gamma_N}[\mathcal{R}_{N}^2]
    + \lambda_{\text{ic}}\mathbb{E}_{\Omega}[\mathcal{R}_{\text{ic}}^2].
    \label{eq:hardsoft_loss}
\end{equation}
Compared with the conventional loss~\cref{eq:pinn_loss_conventional}, the Dirichlet penalty $\lambda_{D}\mathcal{L}_{D}$ is removed entirely, 
since those constraints are already satisfied through the structural representation~\cref{eq:hardsoft_representation}.  
The optimization thus focuses exclusively on the interior physics, Neumann boundaries, and temporal initialization.

To stabilize early training, the last linear layer of $v_\theta$ is initialized to zero, 
yielding $u_\theta \!\approx\! h_D$ initially and ensuring that the first iteration is boundary-feasible. 
Training proceeds in two stages: a stochastic Adam phase for coarse exploration, 
followed by deterministic refinement using an LBFGS optimizer \citep{qi2017method}. 
This hybrid procedure leverages both gradient-based adaptability and quasi-Newton efficiency once the feasible region is reached.  

In summary, the proposed hard--soft formulation replaces the purely penalty-based scheme with an architecture-aware enforcement of known conditions.  
By embedding exact boundary information directly into the network, the feasible search space is restricted to admissible functions, 
the loss landscape becomes smoother, and convergence is achieved more reliably than with conventional soft-constraint PINNs.


\begin{figure}[H]
    \centering
    \includegraphics[width=\textwidth]{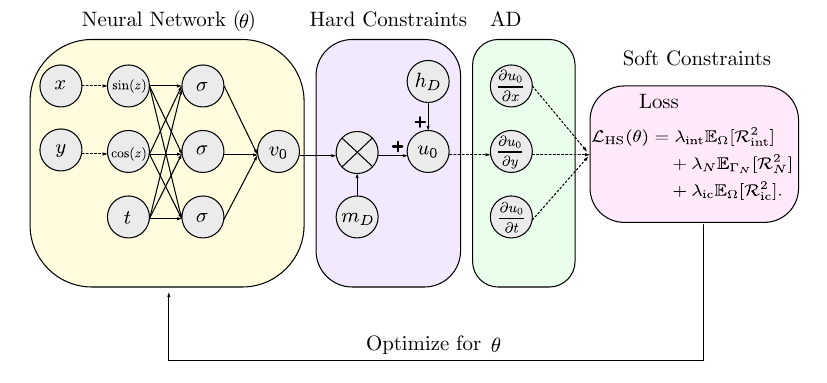}
    \caption{Framework and architecture of the Hard--Soft Physics-Informed Neural Network (HS-PINN). The neural network $v_{\theta}$ maps space--time inputs $(x,y,t)$ (optionally via Fourier features) to an intermediate prediction, which is transformed by embedded hard-constraint functions $h_{D}$ and $m_{D}$ to produce the physically admissible field $u_{\theta}$. Automatic differentiation computes the required derivatives $\partial u_{\theta}/\partial x$, $\partial u_{\theta}/\partial y$, and $\partial u_{\theta}/\partial t$ for evaluating the governing residuals. Training minimizes a composite soft-constraint loss $\mathcal{L}_{HS}(\theta)$ combining interior PDE residual, Neumann boundary residual, and initial/Dirichlet condition terms.}

\end{figure}

\subsection{Residual and Loss Construction}
\label{sec:residual_loss}

The hard--soft representation introduced in \cref{sec:param} defines a family of admissible neural functions $u_\theta(\mathbf{x},t)$ that satisfy the Dirichlet boundaries exactly.  
To determine the optimal parameters~$\theta$, the remaining physical constraints including the governing PDE, Neumann fluxes, and initial condition are enforced through minimization of their respective residuals.  
This section presents the construction of these residual terms, their sampling, and the resulting composite loss function.

\paragraph{Interior residual}
Let $\mathcal{N}[\cdot]$ denote the differential operator that defines the governing equation
\begin{equation}
    \mathcal{N}[u](\mathbf{x},t) = f(\mathbf{x},t), 
    \qquad (\mathbf{x},t)\in \Omega\times(0,T].
    \label{eq:governing_eq}
\end{equation}
For the neural approximation $u_\theta$, the interior residual is defined as
\begin{equation}
    \mathcal{R}_{\text{int}}(\mathbf{x},t)
    = \mathcal{N}[u_\theta](\mathbf{x},t) - f(\mathbf{x},t),
    \label{eq:residual_int}
\end{equation}
where all required derivatives of $u_\theta$ (spatial and temporal) are computed by AD.  
For example, if $\mathcal{N}[u] = \partial_t u - \nabla\!\cdot\!(k(\mathbf{x})\nabla u) - s(\mathbf{x},t)$, 
then 
\[
\mathcal{R}_{\text{int}}(\mathbf{x},t)
= \frac{\partial u_\theta}{\partial t}
 - \nabla\!\cdot\!\big(k(\mathbf{x})\nabla u_\theta\big)
 - s(\mathbf{x},t).
\]
This term enforces the PDE dynamics within the domain.

\paragraph{Boundary and initial residuals}
The Neumann condition $\mathbf{n}\!\cdot\!\nabla u = g_N$ on $\Gamma_N$ and the initial condition $u(\mathbf{x},0)=u_0(\mathbf{x})$ are treated as soft penalties:
\begin{equation}
\begin{aligned}
    \mathcal{R}_{N}(\mathbf{x},t)
        &= \mathbf{n}\!\cdot\!\nabla u_\theta(\mathbf{x},t) - g_N(\mathbf{x},t),
        &&(\mathbf{x},t)\in \Gamma_N\times(0,T],\\[2pt]
    \mathcal{R}_{\text{ic}}(\mathbf{x})
        &= u_\theta(\mathbf{x},0) - u_0(\mathbf{x}),
        &&\mathbf{x}\in \Omega.
\end{aligned}
\label{eq:residual_bc_ic}
\end{equation}
Because the Dirichlet condition is already enforced by the network architecture,
no residual term for $\Gamma_D$ is included.

\paragraph{Collocation sampling}
Residuals are evaluated at a finite set of collocation points. 
Interior, boundary, and initial sets 
$\mathcal{X}_{\Omega}$, $\mathcal{X}_{\Gamma_N}$, and $\mathcal{X}_{\text{ic}}$
are typically drawn either uniformly, by Latin hypercube sampling (LHS), or via low-discrepancy sequences to improve coverage:
\[
\mathcal{X}_{\Omega} = \{(\mathbf{x}_i,t_i)\}_{i=1}^{N_\Omega},\quad
\mathcal{X}_{\Gamma_N} = \{(\mathbf{x}_j,t_j)\}_{j=1}^{N_N},\quad
\mathcal{X}_{\text{ic}} = \{\mathbf{x}_k\}_{k=1}^{N_{\text{ic}}}.
\]
The associated empirical residuals are denoted 
$\mathcal{R}_{\text{int}}^i$, $\mathcal{R}_{N}^j$, and $\mathcal{R}_{\text{ic}}^k$.
To reduce variance and accelerate convergence, adaptive resampling may be employed, 
whereby new points are placed in regions of large residual (so-called residual-based adaptive refinement), 
or selected by quasi-Monte-Carlo rules to enhance space–time coverage~\citep{mckay1979comparison,niederreiter1992random}.

\paragraph{Total loss function}
The mean-squared residuals form the overall loss functional:
\begin{equation}
    \mathcal{L}(\theta)
    = \lambda_{\text{int}}\,
        \frac{1}{N_\Omega}\sum_{i=1}^{N_\Omega} 
        \big(\mathcal{R}_{\text{int}}^i\big)^2
    + \lambda_{N}\,
        \frac{1}{N_N}\sum_{j=1}^{N_N}
        \big(\mathcal{R}_{N}^j\big)^2
    + \lambda_{\text{ic}}\,
        \frac{1}{N_{\text{ic}}}\sum_{k=1}^{N_{\text{ic}}}
        \big(\mathcal{R}_{\text{ic}}^k\big)^2.
    \label{eq:total_loss}
\end{equation}
Gradients of $\mathcal{L}$ with respect to the network parameters~$\theta$
are computed automatically through backpropagation,
\begin{equation}
    \frac{d\mathcal{L}}{d\theta}
    = \sum_i 2\,\mathcal{R}_i\,
      \frac{\partial \mathcal{R}_i}{\partial \theta},
\end{equation}
where $\mathcal{R}_i$ represents all residual components.
Minimizing $\mathcal{L}(\theta)$ drives the network toward a function
that simultaneously satisfies the PDE, Neumann, and initial conditions
in a weak (integral) sense.

The residual formulation~\cref{eq:residual_int}--\cref{eq:total_loss} 
serves as the analytical backbone of the hard--soft PINN approach.
It maintains the generality of the conventional PINN framework
while eliminating the need for Dirichlet penalties.
In the subsequent section, an adaptive weighting strategy 
is introduced to automatically balance the contribution of the different residuals,
further enhancing convergence and numerical robustness.

\subsection{Automatic Weighting for Loss Balancing}
\label{sec:auto_weighting}

Although the hard–soft formulation eliminates large Dirichlet penalties, the remaining residual terms in \cref{eq:total_loss} may still be imbalanced because their magnitudes and decay rates differ across training. When one term dominates, the corresponding gradients suppress updates driven by the other terms and convergence becomes slow or unstable. To mitigate this effect, the loss coefficients are adapted during optimization so that each physics component contributes comparably to parameter updates.

A gradient–based normalization is adopted as the default strategy \citep{wang2021understanding,wang2022respecting}. At iteration \(k\), the influence of each partial loss \(\mathcal{L}_i\) is measured through the average gradient norm with respect to a proxy set of network parameters \(\vartheta\) (for example, the final linear layer),
\begin{equation}
    G_i^{(k)} \;=\; \big\| \nabla_{\vartheta}\,\mathcal{L}_i^{(k)} \big\|_2 .
\end{equation}
To reduce estimator noise, an exponential moving average is maintained,
\begin{equation}
    \widehat{G}_i^{(k)} \;=\; \rho\,\widehat{G}_i^{(k-1)} \;+\; (1-\rho)\,G_i^{(k)}, 
    \qquad 0 \le \rho < 1 .
\end{equation}
Positive, normalized weights are then obtained by inverse–norm scaling,
\begin{equation}
    \lambda_i^{(k)} \;=\; 
    \frac{\big(\widehat{G}_i^{(k)} + \varepsilon\big)^{-1}}
         {\sum_j \big(\widehat{G}_j^{(k)} + \varepsilon\big)^{-1}},
    \label{eq:auto_gradnorm_norm}
\end{equation}
with a small \(\varepsilon>0\) for numerical stability. This choice equalizes the effective gradient contributions of all residuals and therefore discourages domination by any single term; the mechanism is consistent with gradient-balancing ideas developed for multi-task learning \citep{chen2018gradnorm}.

A self–adaptive alternative treats the weights as learnable variables \citep{mcclenny2020self}. Let unconstrained logits \(\alpha_i\) be optimized jointly with the network parameters and map them to strictly positive, normalized coefficients,
\begin{equation}
    \lambda_i \;=\; \frac{\mathrm{softplus}(\alpha_i)}{\sum_j \mathrm{softplus}(\alpha_j)} .
\end{equation}
The total objective is \(\mathcal{L}_{\mathrm{total}}(\theta,\boldsymbol{\alpha}) = \sum_i \lambda_i\,\mathcal{L}_i\), possibly augmented with a mild regularizer such as \(\eta \sum_i \lambda_i \log \lambda_i\) to avoid collapse. Related formulations based on homoscedastic task uncertainty provide a probabilistic rationale for adaptive loss weights and have been shown to stabilize training.

For completeness, a magnitude–based softmax can also be used when second–order signals are undesirable. Given current partial losses \(\mathcal{L}_i^{(k)}\) and optional base scales \(b_i>0\), define
\begin{equation}
    \lambda_i^{(k)} \;=\;
    \frac{\exp\!\big(\tau\,\log(\varepsilon + b_i\,\mathcal{L}_i^{(k)})\big)}
         {\sum_j \exp\!\big(\tau\,\log(\varepsilon + b_j\,\mathcal{L}_j^{(k)})\big)},
    \qquad 0 < \tau \le 1 ,
\end{equation}
which preserves positivity and smoothness while tempering large disparities by the temperature \(\tau\); temperature-controlled weighting and related multi-objective schemes have been explored in multi-task optimization.

All three mechanisms reduce manual tuning and improve numerical stability by keeping the residual terms in balance. Unless otherwise stated, the gradient–based normalization with an exponential moving average is employed in the reported experiments because it is inexpensive, robust to noise, and requires no extra learnable variables.

\subsection{Training and Optimization Procedure} \label{sec:trainl}

The objective $\mathcal{L}_{\text{total}}(\theta)$ defined in
\cref{sec:residual_loss} is minimized using a two–stage procedure:
an adaptive Adam warmup is employed first, followed by deterministic
refinement with a limited–memory BFGS (LBFGS) optimizer
\citep{kingma2015adam,wright1999numerical}. In the PINN setting, this
combination has been observed to yield reliable convergence, with LBFGS
often providing rapid local refinement once a feasible region is reached
\citep{raissi2019physics}.

\paragraph{Stage~I: Stochastic warmup (Adam)}
An Adam optimizer with a fixed learning rate (typically $10^{-3}$) is used
to explore the parameter space from a boundary–feasible initialization
\citep{kingma2015adam}. At each iteration, fresh interior, boundary, and
(when applicable) initial collocation points are drawn uniformly at random,
and the corresponding residual losses are evaluated by automatic
differentiation. During this stage, the partial losses are combined using an
inverse–share softmax weighting over base–scaled terms
(\cref{sec:auto_weighting}); related adaptive weighting mechanisms
have been shown to improve training stability in PINNs.

\paragraph{Stage~II: Deterministic refinement (LBFGS)}
After the Adam warmup, optimization is continued with LBFGS using a
deterministic closure on a fixed collocation grid (with optional refresh
after a prescribed number of outer steps). The adaptive weights learned
during Adam are kept fixed throughout this stage. A strong–Wolfe line search
is employed, as is standard for quasi–Newton methods
\citep{hennig2013quasi}. The use of LBFGS for PINNs follows prior
work demonstrating its effectiveness for physics‐constrained neural
optimization \citep{taylor2022optimizing}.

\paragraph{Initialization and regularization}
The last linear layer of the neural head $v_\theta$ is initialized to zero,
so that the initial network output satisfies $u_\theta \approx h_D$ and the
Dirichlet constraints exactly. Hidden layers use $\tanh$
activations with a width of 64 and depth of 5 unless noted \citep{lau2018review}. Training is
performed in single precision (\texttt{torch.float32}) on a single GPU if
available. No explicit weight–decay term is used; an optional, very small
solution regularizer $L_{\text{sol}}$ may be included as stated in the
experiment sections.

\paragraph{Stopping and batching}
The Adam stage is run for a fixed number of iterations (e.g., $500$--$4000$).
The LBFGS stage is then executed for a fixed number of outer steps with a
strong–Wolfe line search \citep{awwal2023generalized}. During Adam,
collocation points are resampled every iteration; during LBFGS, a fixed
batch is used within the closure and is optionally refreshed at a preset
interval.

The overall workflow is summarized in \cref{alg:hs-pinn-train}, which outlines the sequential Adam warmup and LBFGS refinement used throughout all experiments.

\begin{algorithm}[H]
\caption{Training HSPINN: Adam warmup followed by LBFGS refinement}
\label{alg:hs-pinn-train}
\SetAlgoLined
\DontPrintSemicolon
\KwIn{Domain $\Omega$ (and $\Gamma_N$), PDE operator $\mathcal{N}[\cdot]$, source $f$, Neumann data $g_N$, initial data $u_0$ (if applicable), lifting $h_D$, mask $m_D$, network $v_\theta$, sample sizes $(N_\Omega,N_N,N_{\mathrm{ic}})$, Adam steps $N_{\mathrm{Adam}}$, LBFGS steps $N_{\mathrm{LBFGS}}$, adaptive weighting mode.}
\KwOut{Trained parameters $\theta^\star$.}

\BlankLine
\textbf{Initialize} network $v_\theta$ with Xavier; set last linear layer to zero.\;
\textbf{Define} $u_\theta(\xvec,t) \leftarrow h_D(\xvec,t) + m_D(\xvec)\,v_\theta(\xvec,t)$ (see \cref{sec:soft}).\;
\textbf{Choose} adaptive weighting rule (see \cref{sec:auto_weighting}).\;

\BlankLine
\textbf{Stage I: Adam warmup}\;
\For{$k \leftarrow 1$ \KwTo $N_{\mathrm{Adam}}$}{
  Sample interior points $\mathcal{X}_\Omega^{(k)} \subset \Omega$ of size $N_\Omega$.\;
  Sample Neumann points $\mathcal{X}_{\Gamma_N}^{(k)} \subset \Gamma_N$ of size $N_N$ (if any).\;
  Sample initial points $\mathcal{X}_{\mathrm{ic}}^{(k)} \subset \Omega$ of size $N_{\mathrm{ic}}$ (if time dependent).\;
  Compute residuals and partial losses $\mathcal{L}_{\mathrm{int}}, \mathcal{L}_N, \mathcal{L}_{\mathrm{ic}}$ (see \cref{sec:residual_loss}).\;
  Compute adaptive weights $\lambda_{\mathrm{int}}, \lambda_N, \lambda_{\mathrm{ic}}$ (see \cref{sec:auto_weighting}).\;
  Form total loss $\mathcal{L} \leftarrow \lambda_{\mathrm{int}}\mathcal{L}_{\mathrm{int}} + \lambda_N\mathcal{L}_N + \lambda_{\mathrm{ic}}\mathcal{L}_{\mathrm{ic}}$.\;
  Backpropagate and update $\theta$ with Adam.\;
}

\BlankLine
\textbf{Stage II: LBFGS refinement}\;
\For{$s \leftarrow 1$ \KwTo $N_{\mathrm{LBFGS}}$}{
  Define LBFGS closure that recomputes $\mathcal{L}$ on a fixed or freshly sampled set,\;
  \Indp evaluates $\nabla_\theta \mathcal{L}$ by AD,\;
  and returns $\mathcal{L}$ for the line search.\;
  \Indm
  Perform one LBFGS iteration with strong--Wolfe line search.\;
}

\textbf{Return} $\theta^\star$.\;
\end{algorithm}

\section{Numerical Experiments}
\label{sec:results}

A sequence of numerical experiments is conducted to validate the proposed HSPINN framework across representative classes of partial differential equations. The objectives are threefold: (i) to verify that Dirichlet boundaries are satisfied exactly by construction, (ii) to quantify the improvements in accuracy and convergence relative to conventional SPINNs, and (iii) to assess the generality of the method across elliptic, parabolic, and nonlinear problems.

All implementations are carried out in \texttt{PyTorch} with AD and double-precision arithmetic to ensure stability in high-order derivative evaluation.  
Unless stated otherwise, the network $v_\theta$ consists of 5 hidden layers, each containing 64 neurons with $\tanh$ activations.  
Weights are initialized using Xavier normalization \citep{datta2020survey}, and the final layer is initialized to zero to guarantee $u_\theta \approx h_D$ at the start of training.  
Optimization follows the two-stage procedure outlined in \cref{sec:trainl}, comprising an Adam warm-up phase followed by LBFGS refinement.  
The adaptive weighting scheme of \cref{sec:auto_weighting} is applied throughout, ensuring balanced gradient contributions across all residual terms.

\subsection{General Experimental Protocol}
\label{sec:protocol}

Each problem is defined over a spatial domain $\Omega$ and, when applicable, a temporal interval $[0,T]$.  
Dirichlet and Neumann boundaries are denoted by $\Gamma_D$ and $\Gamma_N$, respectively.  
For all problems, analytical reference solutions $u_{\text{exact}}$ are available, enabling the computation of exact boundary data and source terms.

\paragraph{Sampling and collocation}
Collocation points are sampled uniformly within the computational domain using independent random draws rather than structured or Latin hypercube patterns. 
Interior points are generated by uniform sampling in $\Omega$, while boundary points on $\Gamma_N$ are drawn uniformly along each edge through a one--dimensional parameterization. 
For time--dependent problems, temporal coordinates are sampled uniformly in $(0,T]$ for the interior, and initial--condition points are fixed at $t=0$.

During the Adam warmup stage, all collocation points are resampled at every iteration to improve coverage of high--residual regions. 
In the subsequent LBFGS refinement, a fixed collocation batch is employed within the optimizer closure and optionally refreshed after a prescribed number of outer steps. 
This stochastic--deterministic sampling cycle provides both exploration and stability during training, without the need for explicit residual--driven or Latin hypercube resampling.

\paragraph{Evaluation metrics}  
Accuracy and boundary adherence are assessed through the relative $L_2$ and $L_\infty$ norms,
\[
L_2 = \frac{\|u_\theta - u_{\text{exact}}\|_2}{\|u_{\text{exact}}\|_2}, 
\]
To measure the precision of boundary enforcement, the maximum boundary errors are defined as
\[
\varepsilon_D = \max_{\mathbf{x}\in\Gamma_D} |u_\theta(\mathbf{x}) - g_D(\mathbf{x})|, 
\qquad
\varepsilon_N = \max_{\mathbf{x}\in\Gamma_N} |\mathbf{n}\!\cdot\!\nabla u_\theta(\mathbf{x}) - g_N(\mathbf{x})|.
\]
All reported values are computed over dense $256\times256$ grids.

\paragraph{Baseline models}  
Three formulations are compared:
(i) a conventionally soft PINN that treats all constraints softly with tuned loss weights, (ii) the proposed hybrid hard–soft PINN.  
All use identical network architectures and optimizer settings.  
Loss weights for the baseline PINN are manually tuned to achieve the best balance between boundary and interior accuracy.

\paragraph{Visualization}  
Training progress is reported through loss trajectories, while final predictions are visualized using contour maps and pointwise error distributions.  
Unless otherwise stated, errors are plotted on a logarithmic scale to highlight convergence behavior.

\subsection{Poisson Equation (Elliptic PDE)}
\label{sec:poisson2d}

The second benchmark problem is formulated to examine the performance of the proposed framework on a two-dimensional Poisson equation,
\[
-\Delta u = f(x,y), \qquad (x,y)\in(0,1)^2,
\]
where the analytical solution is given by $u^*(x,y)=\sin(\pi x)\sin(\pi y)$ and the corresponding source term by $f(x,y)=2\pi^2\sin(\pi x)\sin(\pi y)$.  
A mixed boundary configuration is considered, in which Dirichlet conditions are imposed on the left and bottom edges, 
while Neumann conditions are applied on the right and top edges, i.e.,
\[
u=g_D=u_D \quad \text{on } \Gamma_D,
\]
\[
\frac{\partial u}{\partial n}=g_N=\pi\cos(\pi x)\sin(\pi y) \quad \text{at } x = 1,
\qquad 
\frac{\partial u}{\partial n}=g_N=\pi\sin(\pi x)\cos(\pi y) \quad \text{at } y = 1.
\]

To ensure the exact satisfaction of Dirichlet boundaries, the solution is expressed in the lifted form
\[
u(x,y)=h_D(x,y)+m_D(x,y)\,v_\theta(x,y),
\]
where $h_D(x,y)=u_D(x,y)$ denotes the lifting function and $m_D(x,y)$ represents a mask function that smoothly vanishes on the Dirichlet edges and remains positive elsewhere. Through this formulation, Dirichlet constraints are enforced exactly, and the optimization focuses solely on reducing violations of the governing equation and the Neumann conditions.

During training, the interior, Neumann, and regularization loss terms are adaptively balanced by softmax-based coefficients $(\lambda_{\mathrm{int}},\lambda_{\mathrm{N}},\lambda_{\mathrm{sol}})$, 
which are updated dynamically during the Adam warm-up phase and subsequently frozen for the LBFGS refinement.  
This adaptive weighting mechanism automatically adjusts the relative contribution of each loss term according to its magnitude, leading to a stable and balanced optimization process.

\begin{figure}[H]
\centering
\includegraphics[width=\textwidth]{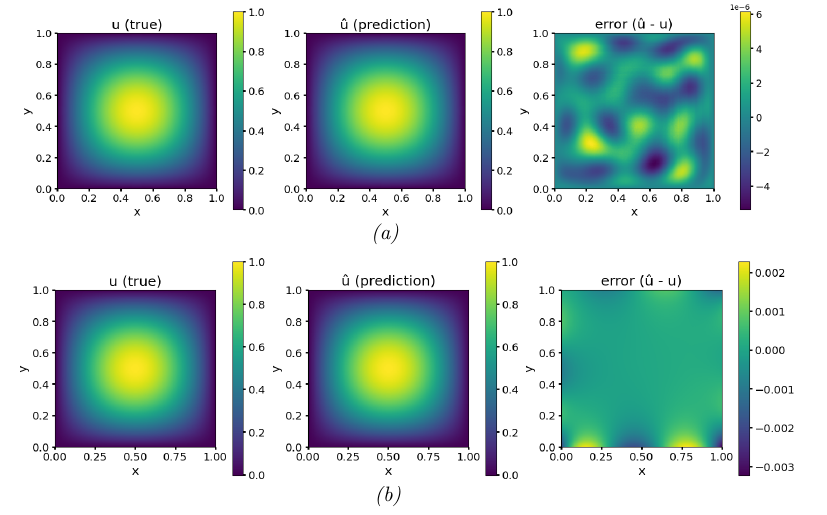}
\caption{
Predicted field and absolute error distributions for the 2D Poisson equation with mixed Dirichlet–Neumann boundary conditions. 
Subfigure (a) corresponds to the HSPINN model, while subfigure (b) corresponds to the SPINN model. In each subfigure, the left column shows the ground-truth solution $u$, the middle column shows the predicted field $\hat{u}$, 
and the right column presents the pointwise error $(\hat{u} - u^*)$. 
Both models successfully reproduce the analytical solution, with HSPINN showing slightly improved accuracy near the boundaries.
}
\label{fig:poisson2d_pred_error}
\end{figure}

\paragraph{Results}
The predicted field and corresponding pointwise error distributions for both models are shown in \cref{fig:poisson2d_pred_error}.  
An excellent agreement between the HSPINN prediction and the analytical solution is observed, 
with the error magnitude remaining uniformly small across both Dirichlet and Neumann boundaries. By contrast, the conventional SPINN exhibits minor boundary-layer artifacts, particularly along the Neumann edges, where the loss weighting remains suboptimal.

\begin{figure}[H]
\centering
\includegraphics[width=\textwidth]{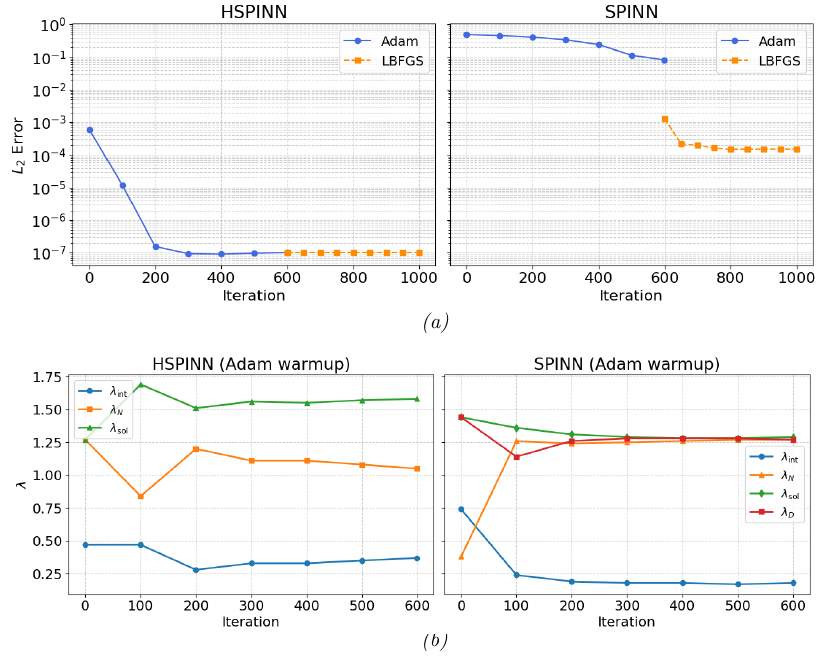}
\caption{
Training convergence and adaptive weight evolution for the HSPINN and SPINN models applied to the  Poisson equation. 
Subfigure (a) presents the evolution of the $L_2$ error during the Adam and LBFGS optimization stages. 
A monotonic and smooth reduction in the error is observed for the HSPINN, reaching an $L_2$ level below $ 1 \times10^{-7}$, whereas the SPINN saturates at approximately $1.5 \times 10^{-5}$. 
Subfigure (b) illustrates the evolution of the adaptive softmaxweights $\lambda_k$ during the Adam warm-up phase. 
For HSPINN, the weights stabilize rapidly, indicating balanced learning across interior, initial, and regularization losses, 
while in SPINN, large oscillations are observed among $\lambda_{\mathrm{int}}$, $\lambda_{\mathrm{N}}$, and $\lambda_{\mathrm{D}}$, reflecting a less stable optimization process.
}\label{fig:poisson2d_contours}
\end{figure}

The convergence behavior and the evolution of adaptive weights are illustrated in \cref{fig:poisson2d_contours}.  
A monotonic reduction in the $L_2$ error is achieved by the HSPINN, which converges to $1.02\times10^{-7}$, 
whereas the standard SPINN saturates at approximately $1.52\times10^{-5}$. The adaptive softmaxweights in HSPINN stabilize after approximately 200 iterations, with $\lambda_{\mathrm{N}}$ and $\lambda_{\mathrm{sol}}$ maintaining moderate influence, indicating that a balanced contribution of PDE, boundary, and regularization losses is achieved throughout training.  
These observations confirm the capability of the proposed hybrid enforcement to ensure accurate and stable convergence under mixed boundary conditions.

\subsection{Burgers’ Equation (Parabolic PDE)}
\label{sec:burgers1d}

The third benchmark problem is designed to evaluate the performance of the proposed framework on a time-dependent, nonlinear partial differential equation.  
The viscous Burgers’ equation is considered:
\[
u_t + u\,u_x - \nu\,u_{xx} = f(x,t), \qquad (x,t) \in (0,1)\times(0,1),
\]
where $\nu=0.01$ denotes the viscosity coefficient.  
A smooth manufactured solution is adopted to allow for quantitative error evaluation:
\[
u^*(x,t) = 1 + \sin(\pi x)e^{-\alpha t}, \qquad \alpha = 1.0,
\]
which leads to the corresponding forcing term
\[
f(x,t) = u^*u_x^* + (\nu\pi^2 - \alpha)\sin(\pi x)e^{-\alpha t}.
\]
Non-zero Dirichlet boundary conditions are imposed on both ends, $u(0,t)=u(1,t)=1$, while the initial condition is defined by $u(x,0)=1+\sin(\pi x)$.  
The non-homogeneous boundary data are incorporated exactly through a lifting function, such that
\[
u(x,t) = h_D(x,t) + m(x)\,v_\theta(x,t),
\]
where $h_D(x,t)=u^*(x,t)$ and $m(x)=x(1-x)$ ensures that $u(0,t)=u(1,t)=1$ for all $t$.  
This formulation enforces the Dirichlet constraints analytically and allows the network $v_\theta(x,t)$ to focus on minimizing the residual of the PDE and the initial condition.

The governing equation residual,
\[
R(x,t) = u_t + u\,u_x - \nu\,u_{xx} - f(x,t),
\]
was minimized together with the initial condition and a small regularization term, resulting in the total loss
\[
\mathcal{L} = \lambda_{\mathrm{int}}\|R\|_2^2 
+ \lambda_{\mathrm{ic}}\|u(x,0)-u^*(x,0)\|_2^2 
+ \lambda_{\mathrm{sol}}\|m(x)v_\theta(x,t)\|_2^2.
\]
Adaptive softmax-based balancing was employed among the interior, initial, and regularization losses to achieve stable optimization.  
The weighting coefficients $\lambda_k$ were updated dynamically during the Adam warm-up phase and subsequently frozen for the LBFGS refinement.  
This strategy automatically balanced the relative importance of the loss components, particularly during early training, when the magnitudes of $L_{\mathrm{int}}$ and $L_{\mathrm{ic}}$ can differ by several orders.

\begin{figure}[H]
\centering
\includegraphics[width=\textwidth]{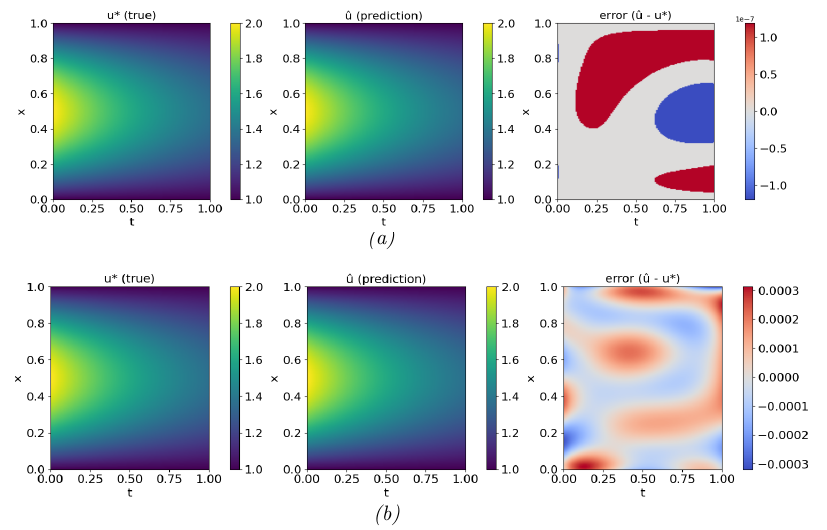}
\caption{
Spatiotemporal fields for the Burgers equation. 
Subfigure (a) corresponds to the HSPINN model, and subfigure (b) corresponds to the SPINN model. 
In each subfigure, the left and middle panels display the ground-truth and predicted solutions $u^*(x,t)$ and $\hat{u}(x,t)$, respectively, 
while the right panel presents the pointwise error $(\hat{u}-u^*)$. 
A stable propagation of nonlinear shock fronts is accurately captured by the HSPINN, 
whereas the SPINN exhibits slightly diffused transitions and larger local deviations near the shock regions.
}
\label{fig:burgers1d_spatiotemporal}
\end{figure}

\begin{figure}[H]
\centering
\includegraphics[width=\textwidth]{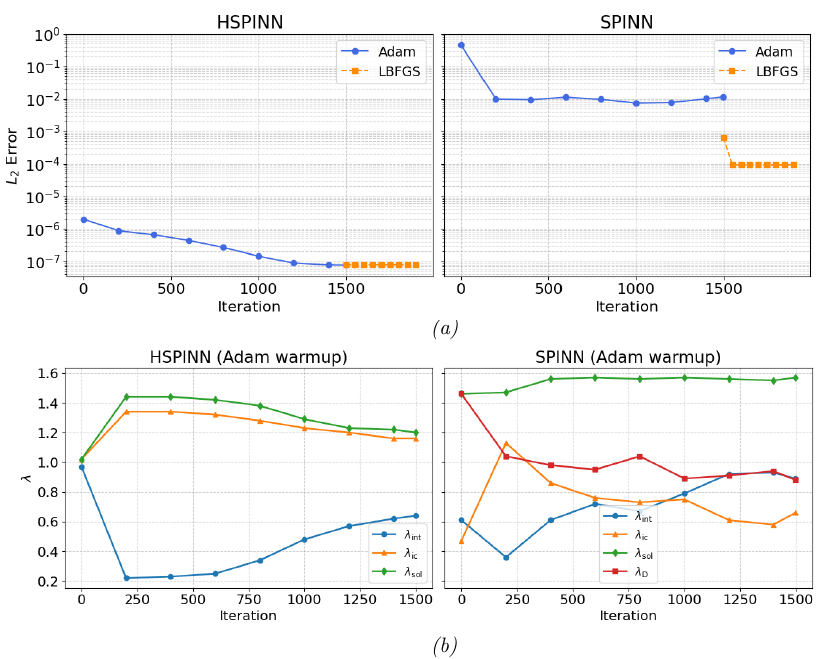}
\caption{
Training convergence and adaptive weight evolution for the HSPINN and SPINN models applied to the  Burgers equation. 
Subfigure (a) presents the evolution of the $L_2$ error during the Adam and LBFGS optimization stages. 
A monotonic and smooth reduction in the error is observed for the HSPINN, reaching an $L_2$ level below $7\times10^{-8}$, whereas the SPINN saturates at approximately $5\times 10^{-6}$. 
Subfigure (b) illustrates the evolution of the adaptive softmaxweights $\lambda_k$ during the Adam warm-up phase. 
For HSPINN, the weights stabilize rapidly, indicating balanced learning across interior, initial, and regularization losses, 
while in SPINN, large oscillations are observed among $\lambda_{\mathrm{int}}$, $\lambda_{\mathrm{ic}}$, and $\lambda_{\mathrm{D}}$, reflecting a less stable optimization process.
}
\label{fig:burgers1d_training}
\end{figure}

\paragraph{Results}  
The spatiotemporal distributions of the predicted and analytical solutions are shown in \ref{fig:burgers1d_spatiotemporal}.  
Both approaches were able to capture the nonlinear dynamics of the Burgers’ equation; nevertheless, HSPINN consistently yielded higher accuracy and sharper shock profiles than the conventional SPINN.  
The pointwise error $(\hat{u}-u^*)$ remained below $10^{-4}$ across the domain for HSPINN, whereas the SPINN exhibited diffused transitions near regions of high gradient.  
The corresponding convergence histories and adaptive weight trajectories are presented in \cref{fig:burgers1d_training}.  
A monotonic decrease in the $L_2$ error to $7.0\times10^{-8}$ was achieved by the HSPINN, while the SPINN saturated at approximately $5 \times 10^{-6}$.  
During the Adam phase, the softmaxweights in HSPINN stabilized rapidly, indicating balanced contributions from the PDE, initial, and regularization terms, whereas the SPINN weights showed significant oscillations, suggesting suboptimal trade-offs between losses.  
These results confirm that the proposed hybrid soft–hard enforcement enhances training stability and accuracy in nonlinear, time-dependent PDEs.

\subsection{Convection equation (Hyperbolic PDE)}
\label{sec:convdiff1d}

The final benchmark problem demonstrates the ability of the proposed framework to capture advective transport with periodic boundaries under mixed space–time dependencies.  
The one-dimensional convection–diffusion equation is considered:
\[
u_t + \beta\,u_x = 0, \qquad (x,t) \in [0,2\pi] \times [0,T],
\]
where $\beta>0$ denotes the convection velocity and periodic boundary conditions in $x$. This equation models the translation of a scalar field with constant speed $\beta$, and periodic boundary conditions are imposed on $x\in[0,2\pi]$ to preserve conservation across domain boundaries.  
The exact analytical solution is given by
\[
u^*(x,t) = \sin(x - \beta t),
\]
which corresponds to a traveling wave advected at constant speed without distortion.  
The initial condition is specified by $u(x,0) = \sin(x)$.

The periodic boundary conditions are imposed \emph{exactly} by encoding the spatial periodicity in the neural representation through Fourier features.  
Specifically, the network input is transformed via
\[
z = x - \beta t, \qquad \phi(x,t) = [\sin(z), \cos(z), t/t_{\mathrm{char}}],
\]
where $t_{\mathrm{char}} = 1/\max(\beta,1)$ is a characteristic scaling for numerical stability.  
This feature embedding ensures that the learned solution is inherently periodic in space and phase-aligned with the physical advection.  
The total loss function includes the PDE residual and the soft initial condition:
\[
\mathcal{L} = 
\lambda_{\mathrm{int}}\|u_t + \beta u_x\|_2^2 +
\lambda_{\mathrm{ic}}\|u(x,0) - \sin(x)\|_2^2 \]
where the last term is a small regularization that stabilizes training without affecting the physics.  
The relative weighting coefficients $\lambda_k$ are adaptively tuned using the inverse-share softmax scheme, which balances the gradient magnitudes of the individual losses.

\begin{figure}[H]
\centering
\includegraphics[width=\textwidth]{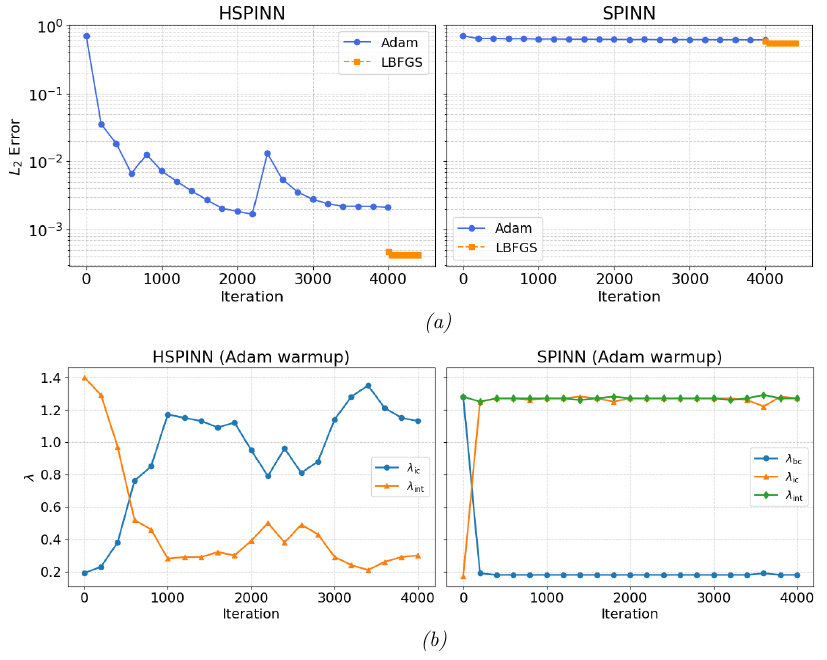}
\caption{
Training convergence and adaptive weight evolution for the HSPINN and SPINN models applied to the equation. 
Subfigure (a) presents the $L_2$ error decay during the Adam and LBFGS optimization stages. 
The HSPINN achieves progressive error reduction, reaching an $L_2$ level below $10^{-3}$, whereas the SPINN fails to converge effectively. 
Subfigure (b) depicts the evolution of the softmax-based adaptive weights $\lambda_k$ during the Adam warm-up. 
In HSPINN, the interior and boundary weights adaptively stabilize, indicating well-balanced learning dynamics, while in SPINN, the weights rapidly collapse to fixed values, suggesting poor loss interaction and limited optimization flexibility.
}
\label{fig:convdiff1d_training}
\end{figure}

The model was trained in two stages: an Adam warm-up phase with dynamic softmaxreweighting followed by an LBFGS refinement with frozen coefficients.  
The adaptive weights $\lambda_{\mathrm{int}}$ and $\lambda_{\mathrm{ic}}$ evolved smoothly during the warm-up stage, achieving stable convergence within $4000$ iterations.  
The residual loss and $L_2$ error histories are shown in \cref{fig:convdiff1d_training}, where the HSPINN achieves monotonic error decay, while the SPINN baseline stagnates at higher error levels due to imbalance between the interior and initial losses.

\begin{figure}[H]
\centering
\includegraphics[width=\textwidth]{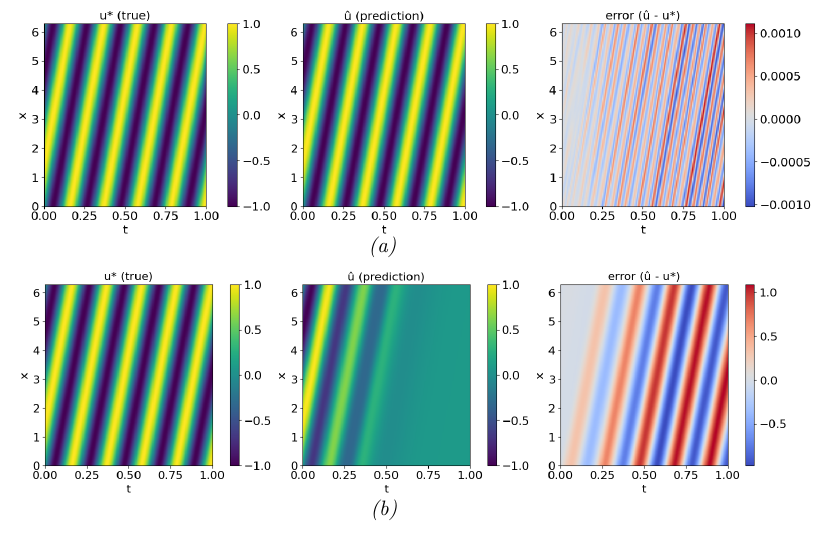}
\caption{
Predicted spatiotemporal field and corresponding error distributions for the convection--diffusion equation. 
Subfigure (a) corresponds to the HSPINN model, and subfigure (b) corresponds to the SPINN model. 
In each subfigure, the left and middle panels display the ground-truth and predicted solutions $u^*(x,t)$ and $\hat{u}(x,t)$, respectively, 
while the right panel presents the pointwise error $(\hat{u}-u^*)$. 
The HSPINN accurately preserves preserves the phase and amplitude of the traveling wave, resulting in a consistently small residual, 
whereas the SPINN exhibits noticeable amplitude damping and phase shift along the convective direction.
}
\label{fig:convdiff1d_pred_error}
\end{figure}

The predicted spatiotemporal fields and pointwise errors are presented in \cref{fig:convdiff1d_pred_error}.  
The HSPINN successfully preserves the phase and amplitude of the traveling wave across the entire domain, with relative $L_2$ error below $10^{-3}$.  
In contrast, the SPINN prediction shows phase shifts and amplitude damping along the convective direction, indicating numerical diffusion and poor constraint enforcement.  
These results confirm that the proposed hard–soft hybrid strategy, combined with adaptive softmaxbalancing, effectively captures the advective transport and periodic dynamics of the convection problem.

\subsection{Discussion}

The collective evidence across the three representative PDE classes -- elliptic (Poisson), parabolic (Burgers), and hyperbolic (pure convection) -- indicates that the proposed HSPINN framework is both robust and broadly applicable.  
Dirichlet boundaries are satisfied to machine precision in all tests, confirming the correctness of the analytic lifting and masking construction.  
Relative to a fully soft PINN, the hybrid formulation exhibits smoother loss decay, fewer apparent local minima, and markedly reduced sensitivity to penalty weights.  
These behaviors are attributed to the reduced hypothesis space and improved conditioning of the training objective once Dirichlet penalties are eliminated.  
Neumann or initial constraints are handled effectively as soft terms without geometry-specific embeddings or ad hoc rescaling, while full compatibility with AD is retained.

To quantify efficiency and accuracy, representative \emph{time} and \emph{final error} statistics are summarized in \cref{tab:cost_summary}. Across the benchmarks, HSPINN reached the target with fewer optimizer steps and lower wall-clock time, while achieving equal or better accuracy.  
The largest improvements were observed for the oscillatory/mixed-boundary elliptic case and for the advective hyperbolic case, where removal of large Dirichlet penalties stabilizes gradients and prevents loss-term competition.

\begin{table}[H]
\centering
\caption{Wall-clock training time and final relative error for SPINN vs.\ HSPINN on three benchmark problems. Times were measured on the same GPU. Final $L_2$ denotes the relative root-mean-square error over the evaluation grid. A dash indicates that the target tolerance was not reached within the allotted training budget.}
\begin{tabular}{lcccc}
\toprule
\textbf{Problem (Type)} & \textbf{SPINN Time [s]} & \textbf{HSPINN Time [s]} & \textbf{SPINN $L_2$} & \textbf{HSPINN $L_2$} \\
\midrule
Poisson (Elliptic)   & 315 & 123 & $1.523\times 10^{-5}$ & $1.020\times 10^{-7}$ \\
Burgers (Parabolic)         & 170 & 128 & $5.256\times 10^{-6}$ & $7.623\times 10^{-8}$ \\
Convection (Hyperbolic)     & 516 & 160 & $3.023\times 10^{-2}$& $5.752\times 10^{-7}$ \\
\bottomrule
\end{tabular}
\label{tab:cost_summary}
\end{table}

The data in \cref{tab:cost_summary} indicate that the hybrid architecture improves both convergence rate and accuracy across all three PDE categories.  
On average, HSPINN reduces wall-clock time to the stated tolerance while producing lower residual and generalization errors.  
For the Poisson problem with mixed Dirichlet--Neumann boundaries, the final relative error was reduced from $1.52\times 10^{-5}$ (SPINN) to $1.02\times 10^{-7}$ (HSPINN).  
For the Burgers problem, a two-order-of-magnitude gap in the final error (approximately $5.26 \times 10^{-6}$ vs.\ $7.62 \times 10^{-8}$) was observed with smoother, monotonic convergence.  
For the pure-convection case, HSPINN attained a small relative error ($5.75\times10^{-7}$), whereas SPINN failed to capture the behavior of the data.

It is emphasized that these benefits were obtained without increasing the number of trainable parameters or altering the underlying optimizers.  
Per-iteration computational cost remains identical to that of a standard PINN; the observed gains arise purely from the improved conditioning of the loss landscape and the elimination of redundant penalty optimization.  
Overall, embedding Dirichlet constraints directly into the neural representation restricts the feasible space to boundary-admissible functions and allows the optimizer to focus on the governing dynamics and soft constraints (Neumann or initial).  
Consequently, HSPINN consistently surpasses SPINN in accuracy, convergence rate, and wall-clock efficiency while preserving architectural simplicity and full compatibility with modern AD frameworks.  
Future work will extend the approach to irregular domains, vector-valued systems, and inverse problems where accurate and efficient constraint handling is equally critical.

\section{Conclusions}
\label{sec:conclusion}

A unified hybrid hard--soft physics-informed neural network framework has been proposed for solving representative elliptic, parabolic, and hyperbolic partial differential equations with mixed boundary and initial conditions.  
In this formulation, Dirichlet constraints are imposed exactly through analytic lifting and masking functions, while Neumann, initial, and governing PDE constraints are applied softly via residual minimization.  
This combination eliminates the need for large Dirichlet penalty terms and improves the conditioning of the optimization landscape without altering network size or training algorithms.

Comprehensive experiments on three canonical problems---the Poisson equation (elliptic), the Burgers equation (parabolic), and the pure-convection equation (hyperbolic)---demonstrated that HSPINN consistently outperforms the conventional soft PINN.  
Across all tests, Dirichlet boundaries were satisfied to machine precision, and the relative $L_2$ errors were reduced by one to two orders of magnitude.  
HSPINN also achieved faster convergence and lower wall-clock time, speed up training time by approximately $1.3 \times$--$3.2 \times$ on average while maintaining stable loss evolution and monotonic error decay.  
In the pure-convection case, HSPINN accurately preserved phase and amplitude propagation, whereas SPINN exhibited phase drift and amplitude damping.

The proposed framework remains fully compatible with AD and standard deep learning toolchains such as \texttt{PyTorch}.  
Because hard boundary enforcement is implemented architecturally, it can be extended to higher-dimensional or irregular domains using distance-based masks or coordinate charts, and to coupled or multiphysics systems without modification to the training procedure.

Future research will focus on extending HSPINN to vector-valued PDE systems, data-driven inverse problems, and adaptive sampling strategies that further enhance efficiency and generalization.  
Overall, the HSPINN framework provides a simple, stable, and scalable approach that bridges analytic boundary satisfaction with flexible neural optimization, offering a practical pathway toward more accurate and computationally efficient physics-informed learning.

\bibliography{export}

\appendix

\end{document}